\documentclass[preprint,12pt]{elsarticle}

\usepackage{amssymb}

\usepackage{amsmath}
\usepackage{bm}
\usepackage{amsfonts}
\usepackage{amssymb}
\usepackage{amsthm}
\usepackage{hyperref}
\usepackage[margin = 1 in]{geometry}
\usepackage{bbm}
\usepackage{graphicx}
\usepackage{listings}
\usepackage[svgnames]{xcolor}
\usepackage{algorithmic}
\usepackage{enumitem}
\usepackage{bbold}
\usepackage{dsfont}
\usepackage{float}
\usepackage[font=footnotesize, labelfont=bf]{caption}
\usepackage[ruled,vlined]{algorithm2e}
\usepackage{array,multirow,graphicx}

\newcommand{\bigbrac}[1]{\left(#1\right)}
\newcommand{\bigmbrac}[1]{\left[#1\right]}
\newcommand{\biglbrac}[1]{\left\{#1\right\}}

\journal{Journal of Biomedical Informatics}

\begin{document}

\begin{frontmatter}



\title{Federated Learning Algorithms for Generalized Mixed-effects Model (GLMM) on Horizontally Partitioned Data from Distributed Sources}

%

\author[inst1]{Wentao Li}
\author[inst2]{Jiayi Tong}
\author[inst3]{Md. Monowar Anjum}
\author[inst3]{Noman Mohammed}
\author[inst2]{Yong Chen}
\author[inst1]{Xiaoqian Jiang\corref{cor1}}
\cortext[cor1]{\url{xiaoqian.jiang@uth.tmc.edu}}
\affiliation[inst1]{organization={School of Biomedical Informatics, UTHealth},
            addressline={7000 Fannin St}, 
            city={Houston},
            postcode={77030}, 
            state={TX},
            country={USA}}


\affiliation[inst2]{organization={Department of Biostatistics, Epidemiology and Informatics, University of Pennsylvania},
            addressline={3400 Civic Center Boulevard}, 
            city={Philadelphia},
            postcode={19104}, 
            state={PA},
            country={USA}}

\affiliation[inst3]{organization={Department of Computer Science, University of Manitoba},
            city={Winnipeg},
            country={Canada}}

\begin{abstract}
\textit{Objectives}: This paper \textcolor{black}{developed federated solutions based on two approximation algorithms} to achieve federated generalized linear mixed effect models (GLMM). \textcolor{black}{The paper also proposed a solution for numerical errors and singularity issues.} And compared the developed model's outcomes with each other, as well as that from the standard R package (`lme4'). 

\noindent\textit{Methods}: The log-likelihood function of GLMM is approximated by two numerical methods (Laplace approximation and Gaussian Hermite approximation), which supports federated decomposition of GLMM to bring computation to data. \textcolor{black}{To solve the numerical errors and singularity issues, the loss-less estimation of log-sum-exponential trick and the adaptive regularization strategy was used to tackle the problems caused by federated settings.}

\noindent\textit{Results:} Our \textcolor{black}{proposed} method can handle GLMM to accommodate hierarchical data with multiple non-independent levels of observations in a federated setting. The experiment results demonstrate comparable (Laplace) and superior (Gaussian-Hermite) performances with simulated and real-world data. 

\noindent\textit{Conclusion:} We \textcolor{black}{modified} and compared federated GLMMs with different approximations, which can support researchers in analyzing \textcolor{black}{versatile} biomedical data to accommodate mixed effects and address non-independence due to hierarchical structures (i.e., institutes, region, country, etc.). 


\end{abstract}




\end{frontmatter}


\section{Introduction}

There is an increasing surge of interest in analyzing biomedical data to improve health. Biostatisticians and machine learning researchers are keen to access personal health information for a deeper understanding of diagnostics, disease development, and potential preventive or treatment options \cite{telenti2020treating}.

In the US, healthcare and clinical data are often collected by local institutions. For many situations, combining these datasets would increase statistical power in hypothesis testing and provide better means to investigate regional differences and subpopulation bias (e.g., due to differences in disease prevalence or social determinants). However, such an information harmonization process needs to respect the privacy of individuals, as healthcare data contain sensitive information about personal characteristics and health conditions. As a minimum requirement, HIPAA (Health Insurance Portability and Accountability Act)\cite{hippa} specifies PHIs (protected health information) and regulations to de-identify the sensitive information (i.e., safe harbor mechanism). But HIPAA compliance does not mean full protection of the data, as several studies demonstrated re-identifiability of HIPAA de-identified data \cite{bonomi2018linking, janmey2018re, sweeney2017re}. Ethical healthcare data sharing and analysis should also respect the “minimum necessary” principle to  reduce the unnecessary risk of potential data leakage, which might increase the likelihood of information leakage.
\begin{figure}[H]
    \centering
    \includegraphics[scale=0.4]{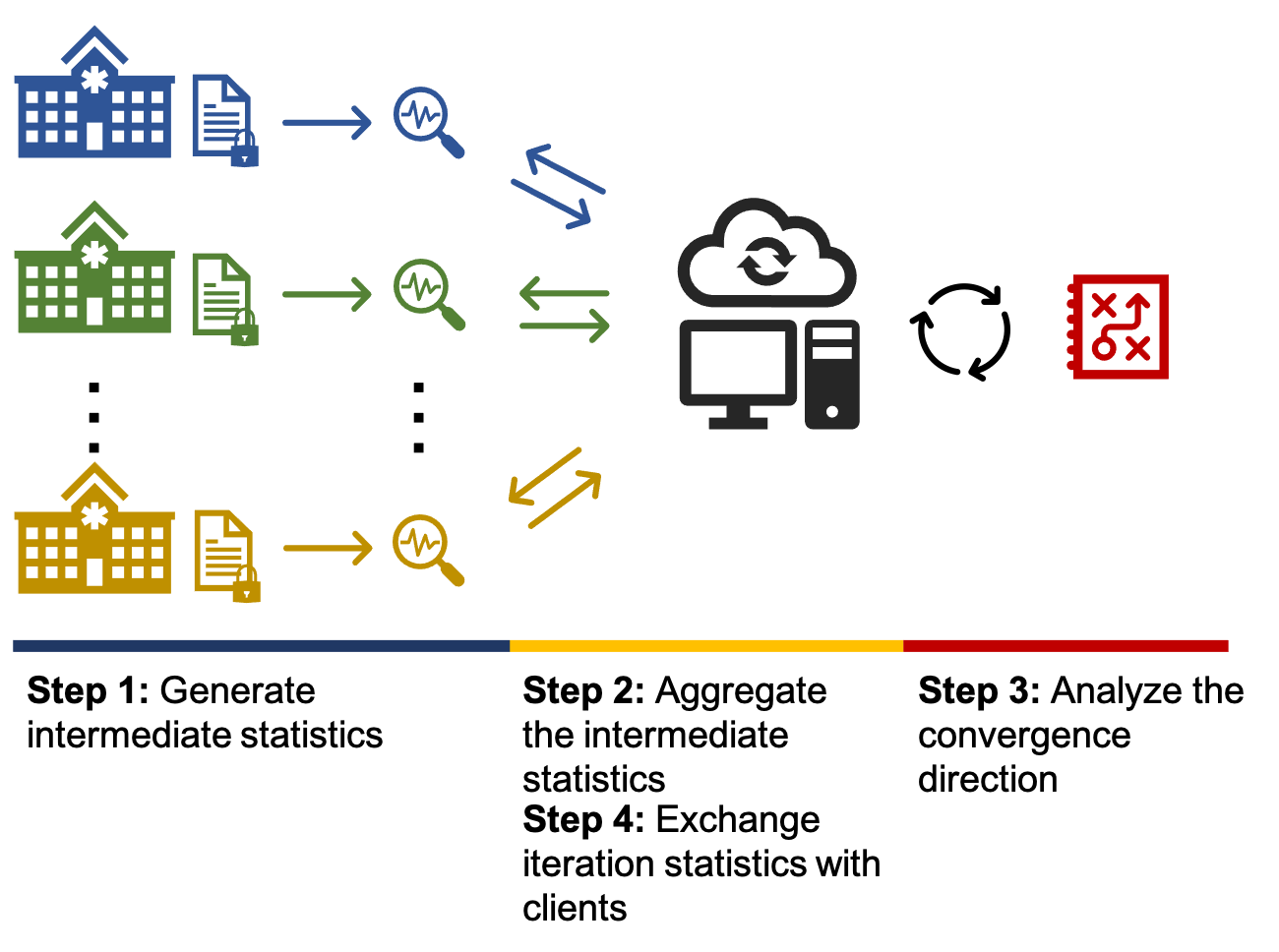}
    \caption{\textbf{Schema of federated learning model in multiple geographically distributed healthcare institutions.} The local institutions periodically exchange intermediate statistics and update the convergence situation of the global model.}
    \label{federated}
\end{figure}

The recent development of federated learning, which intends to build a shared global model without moving local data from their host institutions (Fig. \ref{federated}), shows good promise in addressing the challenge in data sharing mentioned above. Despite the exciting progress, there is still an important limitation as existing models cannot effectively handle mixed-effects (i.e., both fixed and random effects), which is very important to analyzing non-independent, multilevel/hierarchical, longitudinal, or correlated data. \textcolor{black}{Also, due to the sampling errors (i.e., smaller sample size in local sites), variances from these local statistics are larger than those of the global model. These issues, if not addressed appropriately, would lead to failure in global optimization.} The goal of this paper is to improve existing techniques and provide practical solutions with open-source implementation and to allow ordinary biomedical/healthcare researchers to build federated mixed effect learning models for their studies.

\section{Related Work}
Federated learning for healthcare data analysis is not a new topic, and there have been many previous studies. However, most of the existing methods assume the observations are independent and identically distributed\cite{glm, regression}. 
In the presence of non-independence due to hierarchical structures (e.g., due to institutional or regional differences), existing federated models have strong limitations in ignoring the regional differences. The generalized linear mixed model (GLMM), which takes the heterogeneous factors into consideration, is more amenable to accommodate the heterogeneity 
across healthcare systems. There have been very few studies in this area and one relevant work is a privacy-preserving Bayesian GLMM model\cite{zhu2020privacy}, which proposed an Expectation-Maximization (EM) algorithm to fit the model collaboratively on horizontally partitioned data. The convergence process is relatively slow (due to the Metropolis–Hastings sampling in the E-step) and it is also not very stable (likely to be trapped in local optima\cite{xu1996convergence} in high-dimensional data). In the experiment, a loose threshold  (i.e., 0.08) was used as a convergence condition  \cite{zhu2020privacy} while typical federated learning algorithms \cite{glore} in healthcare use much stringent convergence threshold (i.e., $10^{-6}$).

Another related work to fit GLMM in a federated manner is the distributed penalized quasi-likelihood (dPQL) algorithm\cite{luo2021dpql}. This algorithm reduces the computational complexity by considering the target function of penalized quasi-likelihood, which is motivated from Laplacian approximation. The model has communication efficiency over the EM approach and can converge in a few shots. However, the target function PQL can have first order asymptotic bias \cite{lin1996bias} due to the Laplacian approximation of the integrated likelihood.





There is an alternative strategy, Gauss-Hermite (GH), which supports high-order approximation. It is computationally more intensive and requires special techniques to handle the numerical instability of the logSumExp operation (due to the overflow issue when the dimensionality grows in the sum of the exponential terms). We will explain both models in this manuscript and compare their performance on simulated and real-world data.

\section{Methods}
In this section, we will discuss the statistic model along with challenges to be tackled. A high-level schema of the method is shown in algorithm \ref{alg1}.
\subsection{Notation}
Before we introduce the formation of GLMM, let us define some notations.

\begin{table}[H]
\centering
\begin{tabular}{rl|rl}
$i$            & Index of sites                       & $l_i$ & Log-likelihood function for site $i$                                                                     \\
$j$            & Index of patients in a specific site & $\bm\beta$        & Parameters of fixed effect                                                                               \\
$k$            & Index of Hermite polynomial          & $\mu_i$          & Parameters of random effect in site $i$                                                                             \\
$K$            & Order of Hermite polynomial         & $\tau$         & Hyper-parameters                                                                                         \\
$m$            & Number of sites                      & $\bm\theta$       & Parameter space $(\bm\beta,\tau)$                                                                           \\
$n_i$          & Number of patients in site $i$       & $X_{ij}$       & \begin{tabular}[c]{@{}l@{}}A vector represents the data of $j$-th \\ patient in $i$-th site\end{tabular} \\
$\mathcal L_i$ & Likelihood function for site $i$     & $y_{ij}$            & The outcome of patient $j$ from site $i$ \\
$\lambda$ & The parameter of regularization term & $p$ & Number of variables
\end{tabular}
\end{table}
\subsection{Fitting GLMM with quasi-likelihood}
 Let us provide the formation of the GLMM. Define $\mathbb P$ is the distribution of interest and depending on patient-level data $X_{ij},\ y_{ij}$. Define $\phi$ as the distribution of random effects. We can compose the joint distribution as following
\[
\prod_{j=1}^{n_i}\mathbb P(\bm\theta|X_{ij},y_{ij})\phi(\mu_{i};\tau)
\]

Now we have the log-likelihood function of the joint distribution:
\begin{equation}\label{problem}
\log\{\mathcal L(\bm\theta)\}=\sum_{i=1}^{m}\log\biglbrac{\int_{\mu_{i}}\bigmbrac{\prod_{j=1}^{n_i}\mathbb P(\bm\theta|X_{ij},y_{ij})}\phi(\mu_{i};\tau)\text{d}\mu_{i}}
\end{equation}

From the log-likelihood function Eq. \eqref{problem}, one can see that it does not support direct linear decomposition.
In order to support federated learning, we will leverage approximation strategies to make the objective linearly decomposable with simple summary statistics.

We will compare Laplace approximation and Gauss-Hermite approximation in the following sections.

\subsection{Laplace (LA) approximation}
\textcolor{black}{
With the help of Laplace approximation, the integration from Eq.\eqref{problem} can be approximated by an exponential family expression.
\begin{equation}\label{eq1}
\int_{\mu_{i}}f_{\theta}(\mu_{i})\text{d}\mu_{i}=\int_{\mu_{i}}e^{\log{f_{\bm\theta}(\mu_{i})}}\text{d}\mu_{i}\triangleq\int_{\mu_{i}}e^{g(\mu_{i},\theta)}\text{d}\mu_{i}	
\end{equation}
After the deduction (\ref{laplace_approx}), the intractable problem is solved and the objective is to maximize the following formula with respect to $\theta$, where $g$ is an exponential family function defined above (Eq.\eqref{eq1})
}


\[
\sum_{i=1}^{n_i}\bigbrac{g(\hat\mu_{i},\bm\theta)-\dfrac{n_i}{2}\log\bigbrac{g_{\mu\mu}(\hat\mu_{i},\bm\theta)})}
\],
for which the terms are linearly decomposable from local sites. Site $i$ needs to calculate the following aggregated data:
\begin{itemize}
    \item $p\times p$ matrix:
    \begin{equation}\label{la_hessian}
        \dfrac{\hat\omega_{\beta\beta}\hat\omega-\hat\omega_\beta\hat\omega_{\beta}}{\hat\omega^2}+\hat\mu_{\beta\beta}g_\mu+\hat\mu_\beta(\hat\mu_{\beta}g_{\mu\mu}+g_{\mu\beta})+\hat\mu_{\beta}g_{\mu\beta}+g_{\beta\beta}
    \end{equation}
    \item $p$ - dim vector:
    \begin{equation}\label{la_score}
        \dfrac{\hat\omega_\beta}{\hat\omega}+\hat\omega^2g_{\mu\beta}(\hat\mu_i)g_\mu+g_\beta
    \end{equation}
    \item scalar of random effect: $\hat\mu_i$ and first order derivative of $\tau$ by
    \[
        \dfrac{\hat\omega_\tau}{\hat\omega}+\hat\omega^2g_{\mu\tau}(\hat\mu_i)g_\mu+g_\tau
    \]
\end{itemize}
where $\hat\omega=\sqrt{-\dfrac{1}{g_{\mu\mu}(\hat\mu_{i0})}}$
\subsection{Gauss-Hermite (GH) approximation}
Gauss-Hermite approximation \cite{liu1994note} implements Hermite interpolation concerning Eq. \eqref{eq1}. 
\textcolor{black}{
And after the deduction in \ref{guass_hermite_approx}, notice that when the order of Hermite polynomial $K=1$, the objective function is identical to the method with Laplace approximation. Because GH is more generalizable, we will describe the distributed federated learning model on the GLMM problem with the formation of Gauss-Hermite approximation Eq. \eqref{GHA}. For each site $i$, the followings need to calculate and transmit:
}



\begin{itemize}
    \item $p\times p$ matrix:
    \begin{equation}\label{gh_hessian}
        \dfrac{\hat\omega_{\beta\beta}\hat\omega-\hat\omega_\beta\hat\omega_{\beta}}{\hat\omega^2}+\dfrac{1}{\sum_{k=1}^Kf_k}\sum_{k=1}^K\dfrac{\partial}{\partial\bm\beta}(f_{k_\mu}\hat\mu_\beta+f_{k_\omega}\hat\omega_\beta+f_{k_\beta})-\dfrac{1}{(\sum_{k=1}^Kf_k)^2}\|\sum_{k=1}^l(f_{k_\mu}\hat\mu_\beta+f_{k_\omega}\hat\omega_\beta+f_{k_\beta})\|^2_2
    \end{equation}
    \item $p$ - dim vector:
    \begin{equation}\label{gh_score}
        \dfrac{\hat\omega_\beta}{\hat\omega}+\dfrac{1}{\sum_{k=1}^Kf_k}\sum_{k=1}^K(f_{k_\mu}\hat\mu_\beta+f_{k_\omega}\hat\omega_\beta+f_{k_\beta})
    \end{equation}
    \item scalar of random effect: $\hat\mu_i$ and first order derivative of $\tau$ by
    \[
      \dfrac{\hat\omega_\tau}{\hat\omega}+\dfrac{1}{\sum_{k=1}^Kf_k}\sum_{k=1}^K(f_{k_\mu}\hat\mu_\tau+f_{k_\omega}\hat\omega_\tau+f_{k_\tau})
    \]
\end{itemize}

\subsection{\textcolor{black}{Training} Penalization GLMM with GH approximation}
The convergence of the approximation of the likelihood function may be compromised due to over-fitting. Also, for those spatially correlated data, the convergence of them may lead to a complex model. Hence, $L2$ regularization is added to the local log-likelihood function of Gauss-Hermite approximation form, and as shown below
\begin{equation}\label{local_max}
	 l_i=\log\mathcal L_i=\log\bigbrac{\sqrt{2\pi}\hat\omega\sum_{k=1}^Kh_k\exp\biglbrac{g(\hat\mu_{i}+\sqrt{2\pi}\hat\omega x_k;\bm\theta)+x_k^2}} - \lambda\|\bm\beta\|^2_2
\end{equation}
note that when $K=1$, it is represented as regularized Laplace approximation to the problem. To evaluate and find the optimum $\lambda$, we steadily increased the value of $\lambda$ in range $[0,10]$ by $1$. Set $\lambda_{\text{opt}}$ as the optimized regularization term with largest $\sum_i^ml_i$. And choose $\hat{\bm\beta}_{\text{opt}}$ as the optimized estimator for $\bm\beta$.

Due to the limited computation digits, computers are not able to calculate the correct results of the local log-likelihood function $l_i$ of the Gauss-Hermite approximation form as stated above. Such problem is also known as the Log-Sum-Exponential problem and can be solved by shifting the center of the exponential sum for easier computation,
\[
\log\sum_{k=1}^K\exp\biglbrac{g(\hat\mu_{i}+\sqrt{2\pi}\hat\omega x_k;\bm\theta)+x_k^2}=a+\log\sum_{k=1}^K\exp\biglbrac{g(\hat\mu_{i}+\sqrt{2\pi}\hat\omega x_k;\bm\theta)+x_k^2-a}
\]
where $a$ is an arbitrary number.

Thus, the global problem of maximizing $\sum_i^ml_i$ can be divided into several local maximization problems \eqref{local_max}. Each local site $i$ will update the regression intermediates, and they will be combined to update the iteration status. 
Specifically, in each iteration of the federated GLMM algorithm, the following statistics are exchanged from each site to contribute aggregated data for the global model
\begin{table}[H]
\centering
\begin{tabular}{ll}
\hline
\multicolumn{1}{c}{LA} & \multicolumn{1}{c}{GH} \\ \hline
 number of variables $p$  & number of variables $p$   \\
 $p\times p$ matrix (Eq. \ref{la_hessian})  & $p\times p$ matrix (Eq. \ref{gh_hessian})   \\
 $p$ - dim vector (Eq. \ref{la_score}) & $p$ - dim vector (Eq. \ref{gh_score})   \\
 $p$ - dim vector $\bm\beta$  & $p$ - dim vector $\bm\beta$   \\
 scalar $\lambda$  & scalar $\lambda$   \\
 scalar $\hat\mu_i$  & scalar $\hat\mu_i$   \\
\textcolor{black}{scalar first order derivative $\tau$}  & \textcolor{black}{scalar first order derivative $\tau$}   \\
  &  \textcolor{black}{ scalar $K$}  \\ \hline
\end{tabular}
\end{table}

Detailed derivatives with the logistic regression setting of the optimization are presented in the appendix \ref{optimizaiton}.

\begin{algorithm}[H]\label{alg1}
\SetAlgoLined
\KwData{Local data $X_i$ from site $i$}
\KwResult{Global model with coefficients $\hat{\bm\beta}_\text{global}$, test P-values, upper and lower bound}
 Initialization: coefficients $\bm\beta$, random effects $\mu_i$ and $\tau$, regularization term $\lambda$\;
 
 \For{$\lambda=0$ to $10$}{
 	\While{$\Delta\mu\neq 0$}{
 		Maximized $\mu_i$ with respect to $\bm\theta$
 		
 		\While{$\Delta\bm\theta\neq 0$}{
 			1. Approximate the log-likelihood functions $l_\text{approx}$ with Laplace or Gauss-Hermite;
 			
 			2. Calculate intermediate statistics. For Laplace, Eq. \ref{la_hessian}, \ref{la_score}; for Gauss-Hermite, Eq. \ref{gh_hessian}, \ref{gh_score};
 			
 			3. Send the intermediate statistics to center server, and then the center server will aggregate them;
 			
 			4. Update $\bm\theta$ and $l_\text{approx}({\bm\theta},\lambda)$ in center server and send back to each client $i$;
 		}
 	}
 }
 \textbf{Return:} The largest $l_{\text{approx}}(\hat{\bm\theta},\hat\lambda)$, the coefficients $\hat{\bm\beta}_\text{global}=\hat{\bm\beta}$, \textcolor{black}{hyperparameter $\hat\tau$}, and the regularization term $\hat\lambda$
 \caption{Distributed GLMM with approximation methods}
\end{algorithm}

\section{Experiments}
\textcolor{black}{Our algorithm is developed in Python with packages \textit{pandas}, \textit{numpy}, \textit{scipy}, and the benchmark algorithm is glmer function in R package `lme4'.}
\subsection{Benchmarking the methods using synthetic data}
To test the performance of our proposed methods, we first designed a stress test based on a group of synthetic data, which include 8 different settings (Tab.\ref{summary_data}), and each set contains 20 datasets. In each dataset, it consists of 4 categorical variables with value in $\{0,1\}$; 6 categorical variables with value in range $[-1,1.5]\in\mathbb R$; 1 outcome variable with value in $\{0,1\}$; Site ID, represents the id of which site the entry belongs to; Site sample size, represents the number of samples in this specific setting; Log-odds ratio for each sample; Number of true positive, true negative, \textcolor{black}{true} positive, false positive, false negative.

To evaluate which method can reach better performance, we proposed the following evaluation measurements: discrimination of the estimated coefficients $\hat{\bm\beta}$, the test power of each coefficient, and the precision and recall of the number of significant coefficients.

\begin{table}[H]\label{summary_data}
\caption{{The summary of data in each setting.}}
\centering
\begin{tabular}{cccc}
\hline
\multicolumn{1}{c}{Setting} & \multicolumn{1}{c}{\begin{tabular}[c]{@{}c@{}}Number of \\ sites\end{tabular}} & \multicolumn{1}{c}{\begin{tabular}[c]{@{}c@{}}Sample size \\ in each site\end{tabular}} & \multicolumn{1}{c}{variance} \\ \hline
1                            & 2                                                                              & 500                                                                                     & small                        \\
2                            & 2                                                                              & 500                                                                                     & large                        \\
3                            & 10                                                                             & 500                                                                                     & small                        \\
4                            & 10                                                                             & 500                                                                                     & large                        \\
5                            & 2                                                                              & 30                                                                                      & small                        \\
6                            & 2                                                                              & 30                                                                                      & large                        \\
7                            & 10                                                                             & 30                                                                                      & small                        \\
8                            & 10                                                                             & 30                                                                                      & large                        \\ \hline
\end{tabular}
\label{summary_data}
\end{table}

The valuation experiments were conducted among federated GLMM with Laplace approximation, federated GLMM with Gauss-Hermite approximation, and centralized GLMM (all of the data stored in single host) in the R package. And the stress test will be run in 160 different datasets in 8 different settings as mentioned in Tab.\ref{summary_data}. \textcolor{black}{All of the data in different settings were randomly separated into training sets and validation sets with a ratio of 7:3. And we trained the federated learning model on training data sets, then by slowly increasing the regularization term $\lambda$, we chose the optimum model with the best Akaike information criterion and Bayesian information criterion performance on the validation sets. All testing was performed on 2017 iMac with 16 GB memory, CPU (4.2 GHz Quad-Core Intel Core i7), macOS Big Sur version 11.6, Python 3.8, and R version 3.5.0.}

\begin{figure}[H]
    \centering
    \includegraphics[scale=0.6]{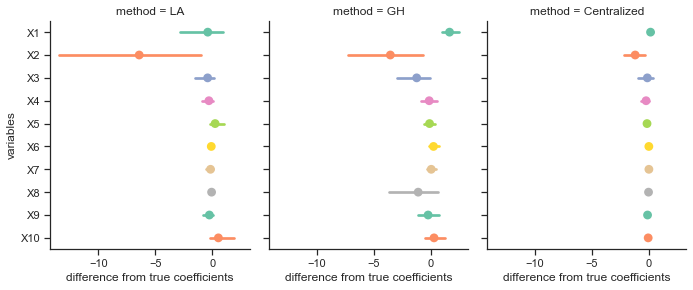}
    \caption{\textbf{The difference from coefficients to the true parameters that are used to generate data.} (Left) The distributed GLMM with Laplace approximation; (Middle) The distributed GLMM with 2-degree Gauss-Hermite approximation. Reminds that $X_1$ is the intercept; (Right) The benchmark of centralized GLMM in R package. }
    \label{coef}
\end{figure}

\begin{figure}[H]
    \centering
    \includegraphics[scale=0.48]{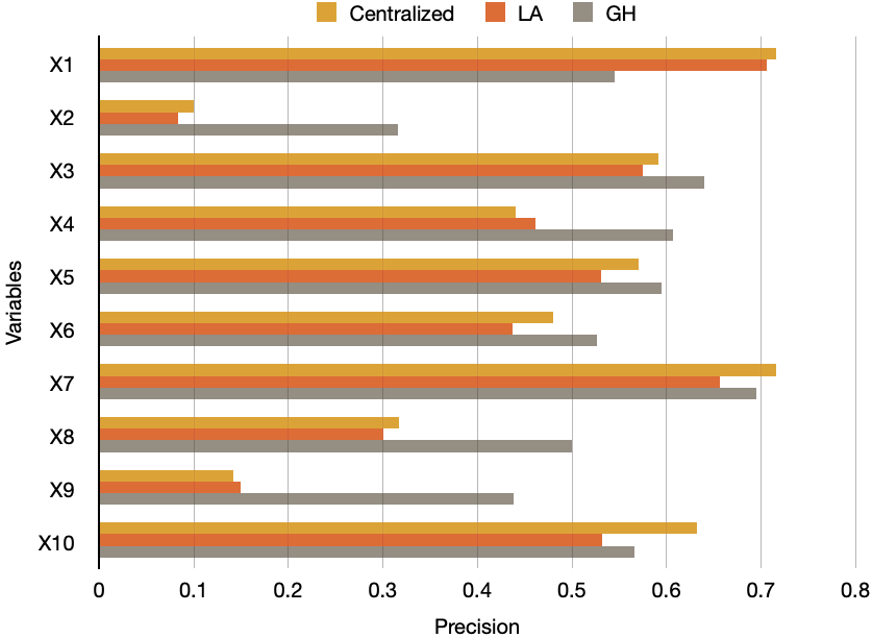}
    \includegraphics[scale=0.48]{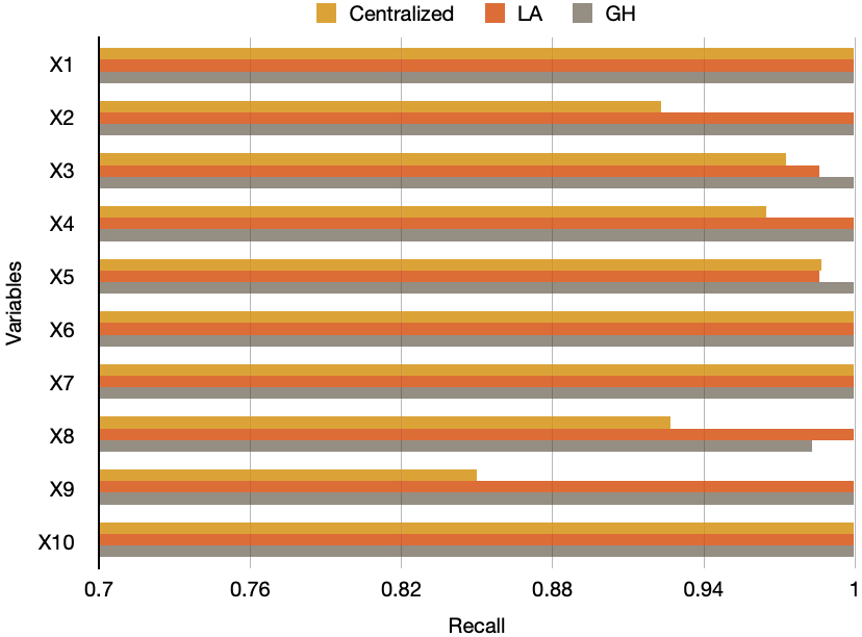}
    \caption{\textbf{The precision and recall among centralized, Laplace, and Gauss-Hermite method under significance level $\alpha=0.05$.} (Left) The precision of the test compared to the true value. (Right) The recall of the test compared to the true value. }
    \label{precision}
\end{figure}

Although we tested the data sets with the state-of-art benchmark algorithm for centralized GLMM in R, the regression is not perfect for the ground truth coefficients we used to generate the data (Fig.\ref{coef}). So, it is also important to have the P-values of variables into consideration when interpreting the model. 
Thus, We made comparisons among centralized GLMM, Laplace method, and Gauss-Hermite method concerning the p-values of coefficients. Tables in the appendix captured the performance of different methods. Fig.\ref{precision} shows the precision and recall results of centralized, Laplace, and Gauss-Hermite methods. Noted that we set our Gauss-Hermite approximation to 2-degree. See tables in Appendix (Tab.\ref{R}, \ref{LA}, \ref{GH}).  

\begin{figure}[H]
    \centering
    \includegraphics[scale=0.6]{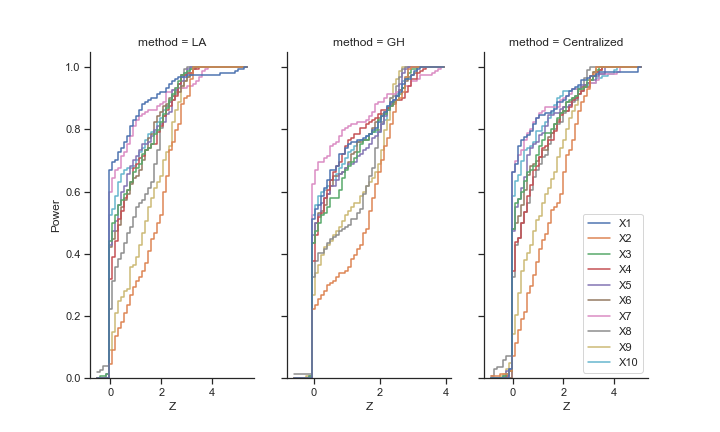}
    \caption{\textbf{The curve of test power among centralized, Laplace, and Gauss-Hermite methods.} (Left) The power of the test of the Laplace method. (Middle) The power of the test of the 2-degree Gauss-Hermite method. (Right) The power of the test of the Centralized method. Power was calculated as the two-sided t-test on p-values among different methods.}
    \label{power}
\end{figure}
\begin{table}[h]
\centering
\caption{\color{black}\textbf{The convergence rates on approximation methods LA and GH.} (Both LA and GH held the same convergence threshold $10^{-3}$. The mean values and standard deviations (in parentheses) were given)}
\color{black}\begin{tabular}{c|cc|cc}\hline
        & \multicolumn{2}{|c|}{LA}                & \multicolumn{2}{c}{GH}               \\ \hline
Setting & Steps            & Runtime (s) & Steps           & Runtime (s) \\ \hline
1       & 22.875 (21.623)  & 47.953 (20.513)    & 34.850 (9.213)  & 104.460 (10.614)   \\
2       & 21.500 (21.977)  & 40.947 (36.466)    & 35.000 (8.711)  & 100.940 (19.940)   \\
3       & 29.867 (31.719)  & 108.931 (65.486)   & 34.900 (6.138)  & 1259.285 (231.956) \\
4       & 27.846 (24.034)  & 84.343 (76.502)    & 36.650 (6.310)  & 1342.695 (250.603) \\
5       & 59.722 (42.057)  & 10.631 (3.945)     & 33.750 (10.146) & 12.568 (2.116)     \\
6       & 67.188 (48.994)  & 10.499 (4.054)     & 31.400 (11.081) & 11.430 (3.064)     \\
7       & 96.286 (53.635)  & 96.501 (38.632)    & 37.450 (3.818)  & 369.165 (41.998)   \\
8       & 116.083 (46.479) & 91.304 (62.410)    & 37.150 (4.295)  & 309.693 (36.621)  \\ \hline
\end{tabular}
\label{convergence}
\end{table}

The simulation results showed the federated Gauss-Hermite approximation performed better than the method based on Laplace approximation on every variable. Also, the federated Gauss-Hermite method achieved higher test power (Fig.\ref{power}). \textcolor{black}{When considering the convergence rates between the two approximation methods, both showed less convergence efficiency in Setting 7 and 8 (Tab.\ref{convergence}). The result Indicates that more local sites and smaller sample sizes will make the federated GLMM more inefficient to converge. Also, GH approximation method will required more computation time compared with LA approximation.} In sum, one-degree increase of the approximation function in LA with our developed GH method, GH outperformed LA methods for federated GLMM implementation.

\subsection{Mixed-effects logistic regression on mortality for patients with COVID-19}
We analyzed the data of COVID-19 electronic health records collected by Optum$^\circledR$ from February 2020 to January 28, 2021, from a network of healthcare providers. The dataset has been de-identified and based on HIPAA statistical de-identification rules and managed by Optum$^\circledR$ customer data user agreement. In this database, there are 56,898 unique positive tested COVID-19 patients. After removing the patients with missing data, the final cohort contains 4,531 patients who died and the rest population (41,781) survived. The database contains a regional variable with five levels (Midwest, Northwest, South, West, Others/unknown) to provide privacy-preserving area information to indicate where the samples were collected.

We have conducted a GLMM model (considering region-distinct random effect) using this dataset with the following predictors: age, gender, race, ethnicity, Chronic obstructive pulmonary disease (COPD), Congestive heart failure (CHF), Chronic kidney disease (CKD), Multiple sclerosis (MS), Rheumatoid arthritis (RA), LU (other lung diseases), High blood pressure (HTN),  ischemic heart disease (IHD), diabetes (DIAB), Asthma (ASTH), obesity (Obese). Our proposed method with GH approximation performed the best with both the smallest \textcolor{black}{Akaike information criterion} (AIC) and \textcolor{black}{Bayesian information criterion} (BIC) according to the table of the goodness of fit (Tab.\ref{goodness}). \textcolor{black}{And the performance of different methods can be shown in (Tab.\ref{measures}).}

\begin{table}[H]
\centering
\caption{\textbf{Statistics of goodness of fit among different methods}}
\centering
\begin{tabular}{rccc}
\hline
\multicolumn{1}{l}{} & Log-likelihood & AIC     & BIC     \\ \hline
R                    & 13562.9        & 27165.9 & 27340.8 \\
LA                   & -13695.0       & 27428.0 & 27594.1 \\
\textbf{GH}                   & \textbf{-11.8}          & \textbf{61.6}    & \textbf{227.7}   \\ \hline
\end{tabular}
\label{goodness}
\end{table}

\begin{figure}[H]
    \centering
    \includegraphics[scale=0.55]{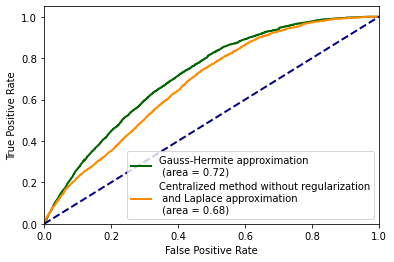}
    \caption{\textbf{The ROC curve with Area Under Curve (AUC) among centralized, Laplace, and Gauss-Hermite methods.} The orange ROC curve is the centralized method without regularization and the Laplace approximation(i.e., R implementation in the `lme4' package, which does not have an option for including regularization). AUC values are also included, a higher AUC value implicates better performance of the model. The green ROC curve is the 2-degree Gauss-Hermite method with regularization.}
    \label{roc}
\end{figure}
\color{black}
\begin{table}[h]
\centering
\caption{\color{black}\textbf{Statistics of performances among different methods}(95\% CIs were generated by Wilson Score interval)}
\color{black}\begin{tabular}{cllllll}
\hline\hline
\multicolumn{1}{l}{}                                                                             &                    & Precision & Recall   & F1-score & AUC      & threshold \\ \hline\hline
\multirow{3}{*}{\begin{tabular}[c]{@{}c@{}}Centralized\\ method\\ with LA\end{tabular}}   & Value              & 0.1507    & 0.6204 & 0.2425   & 0.6789 & 0.0900    \\
                                                                                                 & Lower bound (0.95) & 0.1474    & 0.6160 & 0.2386   & 0.6700 &           \\
                                                                                                 & Upper bound (0.95) & 0.1539    & 0.6248 & 0.2464   & 0.6878 &           \\ \hline
\multirow{3}{*}{\begin{tabular}[c]{@{}c@{}}GH\\ with\\ regularization\end{tabular}} & Value              & 0.1705    & 0.6546 & 0.2705   & 0.7178 & 0.0108    \\
                                                                                                 & Lower bound (0.95) & 0.1670    & 0.6503 & 0.2664   & 0.7091 &           \\
                                                                                                 & Upper bound (0.95) & 0.1739    & 0.6589 & 0.2745   & 0.7265 &           \\ \hline\hline
\end{tabular}
\label{measures}
\end{table}
\color{black}
We also compared the ROC curves (Fig.\ref{roc}) between our proposed GH method and centralized method to check their performance. And the result showed that GH approximation (AUC=0.72) outperforms the centralized method without regularization (AUC=0.68). Indicating GH-based GLMM method has better classification performance than the GLMM based on LA approximation.
In our proposed model, it showed variables: Unknown race, Chronic kidney disease (CKD), Multiple sclerosis (MS), and other lung diseases (LU) are not significant to the mortality of COVID-19. The result of the regression is in the Appendix (Tab.\ref{r_result}, \ref{gh_result}, \ref{la_result}).

\section{Discussion}
\textcolor{black}{We developed solutions to address the limited digit problem (i.e., overflow issue of fixed-length object types due to extremely large numbers in local estimation) using an alternative loss-less estimation of log-sum-exponential term, and the singularity issue (involved in Newton optimization) with an adaptive regularization strategy to avoid inverting low-rank matrices without imposing too much unnecessary smoothness.} 
We \textcolor{black}{further} compared two federated GLMM algorithms \textcolor{black}{with our developed federated solutions} (LA vs. GH) and demonstrated the performance of the federated GLMM based on the GH method surpassed the method based on LA in terms of the accuracy of estimation, power of tests, and AUC. Although the GH method is requiring slightly more computations than the LA method, it is still acceptable for more accurate results. For example, in the prediction of COVID-19 mortality rates, the accuracy of prediction will be more reliable, as we have shown in the previous section. During the optimization iterations, we noticed that some sites have already achieved convergence in very few steps. If those sites stop communicating with the central server, they can be released from extra computations. We would investigate more efficient algorithms based on such a strategy of `lazy regression' for minimizing communication for federated learning models. 

Another limitation of the proposed federated GLMM model is not yet differentially private and iterative summary statistics exchange can lead to incremental information disclosure, which might increase the re-identification risk over time. There are several strategies to improve the model based on secure operations like homomorphic encryption and differential privacy, which we have previously studied in GLM models \cite{Kim2019-gs}. Finally, in practice, there can be extra heterogeneity that cannot be explained by random intercepts only, it is of interest to further develop our algorithms toward GLMM that allows multiple random effects including random coefficients in the regression models.


\section{Acknowledgement}
XJ is CPRIT Scholar in Cancer Research [RR180012]; and he was supported in part by Christopher Sarofim Family Professorship, UT Stars award, UTHealth startup, the National Institute of Health (NIH) under award number [R01AG066749, R01GM114612, U01TR002062]; and the National Science Foundation (NSF) RAPID [\#2027790]. NM is supported by  NSERC Discovery Grants [RGPIN-2015-04147]. JT and YC's research is supported in part by NIH under award number [R01AI130460 and R01LM012607] and Patient-Centered Outcomes Research Institute (PCORI) Project Program Award (ME-2019C3-18315). All statements in this report, including its findings and conclusions, are solely those of the authors and do not necessarily represent the views of the Patient-Centered Outcomes Research Institute (PCORI), its Board of Governors or Methodology Committee.

 \bibliographystyle{elsarticle-num} 
 \bibliography{glmm.bib}

\newpage

\appendix

\section{Supplementary proofs}
\color{black}
\subsection{Laplace approximation}\label{laplace_approx}
Let us explain the Laplace approximation.
Denote that 
\[
f_{\theta}(\mu_{i})\triangleq\prod_{j=1}^{n_i}\mathbb P(\theta|X_{ij},y_{ij})\phi(\mu_{i};\tau)
\]
and one can see the log term inside the log-likelihood function that
\begin{equation}
\int_{\mu_{i}}f_{\theta}(\mu_{i})\text{d}\mu_{i}=\int_{\mu_{i}}e^{\log{f_{\theta}(\mu_{i})}}\text{d}\mu_{i}\triangleq\int_{\mu_{i}}e^{g(\mu_{i},\theta)}\text{d}\mu_{i}	
\end{equation}

 Applying a Taylor expansion on $g(\mu_{i},\theta)$, and we choose $\hat\mu_{i}$ that maximized $g(\mu_{i},\theta)$. See that $\hat\mu_{i}$ satisfies $g_\mu(\hat\mu_{i},\theta)=0$ and $g_{\mu\mu}(\hat\mu_{i},\theta)<0$, we have
 \[
 g(\mu_{i},\theta)= g(\hat\mu_{i},\theta)-\dfrac{1}{2}(\hat\mu_{i}-\mu_{i})^2\bigbrac{-g_{\mu\mu}(\hat\mu_{i},\theta)} + o(\mu_{i}^2)
 \]
 , which is plugged into Eq. \eqref{eq1}. With Laplace approximation \cite{brostrom2011generalized}, one can see that
\begin{align*}
\int_{\mu_{i}}e^{g(\mu_{i},\theta)}\text{d}\mu_{i}\approx exp\{{g(\hat\mu_{i},\theta)}\}\bigmbrac{2\pi\cdot-\dfrac{1}{g_{\mu\mu}(\hat\mu_{i},\theta)}}^{n_i/2}
\end{align*}

\subsection{Gauss Hermite approximation}\label{guass_hermite_approx}
The Hermite polynomial $H_k(x)$ and weight $h_k$ are defined as followings,
\[
H_k(x)\triangleq(-1)^ke^{x^2}\dfrac{d^k}{dx^k}e^{-x^2}\quad\quad
h_k\triangleq\dfrac{2^{k-1}k!\sqrt{\pi}}{k^2\bigmbrac{H_{k-1}(x_k)}^2}
\]
where $x_k$ are the roots of $H_k(x)=0$.

Thus, with the Gauss-Hermite approximation, Eq. \eqref{eq1} can be approximated by
\begin{equation}\label{GHA}
\int_{\mu_{i}}e^{g(\mu_{i})}\text{d}\mu_{i}\approx \sqrt{2\pi}\hat\omega\sum_{k=1}^Kh_k\exp\biglbrac{g(\hat\mu_{i}+\sqrt{2\pi}\hat\omega x_k)+x_k^2},\quad \hat\omega=\sqrt{-\dfrac{1}{g''(\hat\mu_{i})}}
\end{equation}
notice that when \textcolor{black}{$K=1$}, it is a Laplace approximation. 

Our final objective function for GH is
\[
\mathcal L_i=\mathcal L_i(\bm\beta, \mu_{i}; X_{i\cdot}, y_{i\cdot})=\sqrt{2\pi}\hat\omega\sum_{k=1}^Kh_k\exp\biglbrac{g(\hat\mu_{i}+\sqrt{2\pi}\hat\omega x_k;\bm\beta)+x_k^2}
\]

Denote that
\begin{align*}
    &f_k=h_k\exp\biglbrac{g(\hat\mu_{i}+\sqrt{2\pi}\hat\omega x_k;\bm\beta)+x_k^2}\\ 
    &f_{k_\beta}=f_kg_\beta(\hat\mu_i)\\
    &f_{k_\mu}=f_kg_\mu(\hat\mu_{i})\\
    &f_{k_\omega}=f_kg_\mu(\hat\mu_{i})\sqrt{2\pi}x_k
\end{align*}
\color{black}
\subsection{Optimization}\label{optimizaiton}
A logistic regression model with random effects is developed under the form of 
\[
\log\{\mathcal L(\theta)\}=\sum_{i=1}^{m}\log\biglbrac{\int_{\mu_{i}}\bigmbrac{\prod_{j=1}^{n_i}\mathbb P(\theta|X_{ij},y_{ij})}\phi(\mu_{i};\tau)\text{d}\mu_{i}}
\]
and the distribution $\mathbb P$ follows density of logit and $\phi$ is a univariate normal, see that
\[
\prod_{j=1}^{n_i}\mathbb P(\theta|X_{ij},y_{ij})=\prod_{j=1}^{n_i}\pi_{ij}^{y_{ij}}(1-\pi_{ij})^{(1-y_{ij})}
\]
\[
\phi(\mu_{i};\theta)=\dfrac{1}{\sqrt{2\pi}\tau}\exp(-\mu_{i}^2/2\tau^2)
\]
where $\pi_{ij}$ is a Sigmoid function of $\mu_{i}$ and defined as
\[
\pi_{ij}=\dfrac{\exp(X_{ij}^{\top}\bm\beta+\mu_{i})}{1+\exp(X_{ij}^{\top}\bm\beta+\mu_{i})}
\]
with the Gauss-Hermite approximation set up, the objective function can be approximated as
\[
\sqrt{2\pi}\hat\omega\sum_{k=1}^Kh_k\exp\biglbrac{g(\hat\mu_{i}+\sqrt{2\pi}\hat\omega x_k;\bm\beta)+x_k^2}
\]
where
\begin{align*}
	g(\mu_{i};\bm\beta)&=\log\prod_{j=1}^{n_i}\mathbb P(\theta|X_{ij},y_{ij})\phi(\mu_{i};\tau)\\
	&=\sum_{j=1}^{n_i}\bigmbrac{\log\mathbb P(\theta|X_{ij},y_{ij})}+\log\phi(\mu_{i};\tau)\\
	&=\sum_{j=1}^{n_i}\bigmbrac{y_{ij}\log\pi_{ij}+(1-y_{ij})\log(1-\pi_{ij})}+\log\phi(\mu_{i};\tau)
\end{align*}

\subsubsection{Step 1: Maximize $g(\mu_{i})$}
To maximize $g(\mu_{i})$, we need to get the derivatives
\begin{align*}
\dfrac{\partial g}{\partial \mu_{i}}&=\sum_{j=1}^{n_i}\bigmbrac{y_{ij}\dfrac{1}{\pi_{ij}}\dfrac{\partial\pi_{ij}}{\partial\mu_{i}}-(1-y_{ij})\dfrac{1}{1-\pi_{ij}}\dfrac{\partial\pi_{ij}}{\partial\mu_{i}}}+\dfrac{1}{\phi}\dfrac{\partial\phi}{\partial\mu_{i}}\\
&=\sum_{j=1}^{n_i}(y_{ij}-\pi_{ij})-\dfrac{\mu_{i}}{\tau^2}
\end{align*}
where
\[
\dfrac{\partial\phi}{\partial\mu_{i}}=(\sqrt{2\pi}\tau)^{-1}\exp(-\mu_{i}^2/2\tau^2)\cdot(-\mu_{i}/\tau^2)
\]
and
\begin{align*}
	\dfrac{\partial^2 g}{\partial \mu_{i}^2}&=-\sum_{j=1}^{n_i}\dfrac{\partial\pi_{ij}}{\partial\mu_{i}}-\dfrac{1}{\tau^2}<0
\end{align*}
where
\[
\dfrac{\partial\pi_{ij}}{\partial\mu_{i}}=\dfrac{\exp\bigbrac{X_{ij}^\top\bm\beta+\mu_{i}}}{\bigmbrac{1+\exp\bigbrac{X_{ij}^\top\bm\beta+\mu_{i}}}^2}
\]
see that it is a convex problem, using newton's method, we can derive $\hat\mu_{i}=\arg\max_{\mu_{i}}g(\mu_{i})$ that is global optimum.

\subsubsection{Step 2: Maximization preparation of $\beta$ in LOCAL}
See the derivative
\[
\dfrac{\partial\pi_{ij}}{\partial\bm\beta}=\dfrac{X_{ij}\exp\bigbrac{X_{ij}^\top\bm\beta+\mu_{i}}}{\bigmbrac{1+\exp\bigbrac{X_{ij}^\top\bm\beta+\mu_{i}}}^2}
\]
and denote $f_k(\hat\mu_{i};\bm\beta):=h_k\exp\biglbrac{g(\hat\mu_{i}+\sqrt{2\pi}\hat\omega x_k;\bm\beta)+x_k^2}$, then
\begin{align}
	\dfrac{\partial\mathcal L_i}{\partial\bm\beta}&=\sqrt{2\pi}\hat\omega\sum_{k=1}^K\biglbrac{f_k(\hat\mu_{i};\bm\beta)h_k\left.\dfrac{\partial g(\mu_{i};\bm\beta)}{\partial\bm\beta}\right|_{\mu_{i}=\hat\mu_{i}+\sqrt{2\pi}\hat\omega x_k}}\\
	&=\sqrt{2\pi}\hat\omega\sum_{k=1}^K\biglbrac{f_k(\hat\mu_{i};\bm\beta)h_k\sum_{j=1}^{n_i}(X_{ij}y_{ij}-X_{ij}\pi_{ij})}
\end{align}
and the second derivative
\begin{align}
\dfrac{\partial^2\mathcal L_i}{\partial\bm\beta^2}=\sqrt{2\pi}\hat\omega\sum_{k=1}^K&\left\{f_k(\hat\mu_{i};\bm\beta)h_k^2\sum_{j=1}^{n_i}(X_{ij}y_{ij}-X_{ij}\pi_{ij})\bigmbrac{\sum_{j=1}^{n_i}(X_{ij}y_{ij}-X_{ij}\pi_{ij})}^\top\right. \\
&\quad+\left.f_k(\hat\mu_{i};\bm\beta)h_k\sum_{j=1}^{n_i}\bigbrac{-X_{ij}\dfrac{\partial\pi_{ij}}{\partial\bm\beta}}\right\}
\end{align}
Notice that $\mu_{i}$ in (3), (4) and (5) are replaced by $\hat\mu_{i}+\sqrt{2\pi}\hat\omega x_k$ where $\hat\mu_{i}$ is the maximand of function $g(\cdot)$ with respect to $\mu_{i}$.

\subsubsection{Step 3: Maximization of $\beta$ in GLOBAL}
Reminds that $\mathcal L=\sum_{i=1}^m\log\mathcal L_i$, then another Newton's method is applied in global log-likelihood function,
\[
\dfrac{\partial \mathcal L}{\partial\bm\beta}=\sum_{i=1}^m\dfrac{\mathcal L_i'(\bm\beta)}{\mathcal L_i(\bm\beta)}\quad\quad\dfrac{\partial^2 \mathcal L}{\partial\bm\beta^2}=\sum_{i=1}^m\bigmbrac{\dfrac{\mathcal L_i''(\bm\beta)}{\mathcal L_i(\bm\beta)}-\bigbrac{\dfrac{\mathcal L_i'(\bm\beta)}{\mathcal L_i(\bm\beta)}}^2}
\]

Now, focus on $\bm\beta^{(n+1)}=\bm\beta^{(n)}-\dfrac{\mathcal L'(\bm\beta^{(n)})}{\mathcal L''(\bm\beta^{(n)})}$, deduce that
\[
\dfrac{\mathcal L'(\bm\beta^{(n)})}{\mathcal L''(\bm\beta^{(n)})}=\dfrac{\sum_{i=1}^m\dfrac{\mathcal L_i'(\bm\beta)}{\mathcal L_i(\bm\beta)}}{\sum_{i=1}^m\dfrac{\mathcal L_i''(\bm\beta)}{\mathcal L_i(\bm\beta)}-\sum_{i=1}^m\bigbrac{\dfrac{\mathcal L_i'(\bm\beta)}{\mathcal L_i(\bm\beta)}}^2}
\]

\subsection{Synthetic data generation}\label{data_generate}

There are 8 settings of data sets generated from the process, and each setting can be summarized in table \ref{summary_data}.

\begin{figure}[H]
    \centering
    \includegraphics[scale=0.5]{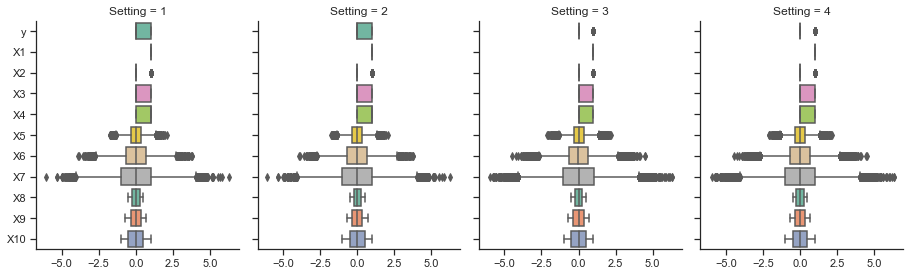}
    \includegraphics[scale=0.5]{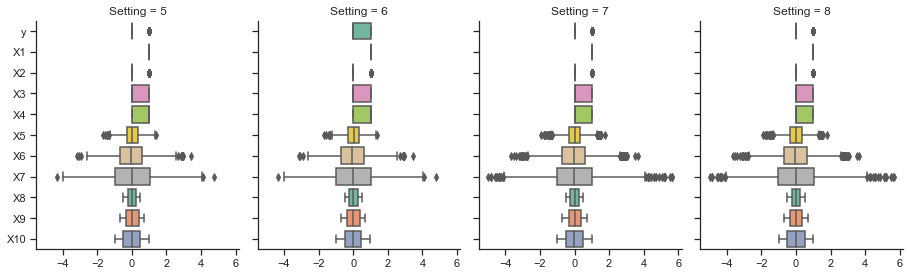}
    \caption{\textbf{The box-plots of variables in Setting 1 to Setting 8.}}
    \label{data}
\end{figure}

Fig.\ref{data} shows the distribution of each setting about different variables. 

We set true sensitivity and specificity as $sen = 0.6\text{ and } sp = 0.9$
and $\bm\beta=(-1.5,0.1,-0.5,-0.3,0.4,-0.2,-0.25,0.35,-0.1,0.5)$. Also define $X_1=\mathbb{1}_N$ as the intercept, and $X_2,X_3,X_4$ are generated with Bernoulli distribution \eqref{eq1} with  probability $p=0.1,0.3,0.5$ respectively.
\begin{equation}\label{eq2}
	f(X_i ; p)=\left\{\begin{array}{ll}
p & \text { if } X_i=1 \\
q=1-p & \text { if } X_i=0
\end{array}\right.
\end{equation}
then $X_5, X_6, X_7$ are generated from normal distributions $\mathcal N(0,0.5),\mathcal N(0,1),\mathcal N(0,1.5)$ respectively. Lastly, $X_8, X_9, X_{10}$ are generate from uniform distributions $\mathcal U(-0.5,0.5)$, $\mathcal U(-0.7,0.7)$, $\mathcal U(-1,1)$ respectively. We also generate the random effect $\mu$ using trivariate normal distribution 
\begin{equation}
	\mathcal N_3\bigbrac{\bigbrac{\begin{array}{ccc}
		0\\0\\0\\
	\end{array}}, \Sigma}, \quad\Sigma=I_3
\end{equation}
and with the settings, we can deduce the log-odds ratio with following formula
\begin{equation}
\log(\pi)=f(X\bm\beta +\mu+\epsilon)
\end{equation}
where $f$ is the sigmoid function defined as
\begin{equation}
f(x)=\dfrac{e^x}{1+e^x}
\end{equation}

Now, we generate the outcomes $y$ for each sample with Bernoulli distribution \eqref{eq1} where the log-odds ratio served as the probability $p$.
Also, the sensitivity and specificity can be calculated with Binomial distribution with probability $sen+\mu_2$ and $sp+\mu_3$.

\section{Supplementary tables}
\begin{table}[H]
\caption{The performance of centralized GLMM on R package with a significance threshold $\alpha=0.05$. TNR refers to the True negative rate.}
\centering
\begin{tabular}{r|llll}
    & \textbf{Precision} & \textbf{Recall} & \textbf{TNR} & \textbf{Accuracy} \\ \hline
X1  & 0.7162                              & 1.0000                           & 0.0000                        & 0.7162                                                                                  \\
X2  & 0.1000                              & 0.9231                           & 0.2000                        & 0.2635                                                                                  \\
X3  & 0.5917                              & 0.9726                           & 0.3467                        & 0.6554                                                                                  \\
X4  & 0.4400                              & 0.9649                           & 0.2308                        & 0.5135                                                                                  \\
X5  & 0.5703                              & 0.9865                           & 0.2568                        & 0.6216                                                                                  \\
X6  & 0.4797                              & 1.0000                           & 0.0000                        & 0.4797                                                                                  \\
X7  & 0.7162                              & 1.0000                           & 0.0000                        & 0.7162                                                                                  \\
X8  & 0.3167                              & 0.9268                           & 0.2336                        & 0.4257                                                                                  \\
X9  & 0.1417                              & 0.8500                           & 0.1953                        & 0.2838                                                                                  \\  
\end{tabular}
\label{R}
\end{table}

\begin{table}[H]
\caption{The performance of distributed GLMM with Laplace transformation with a significance threshold $\alpha=0.05$. TNR refers to the True negative rate.}
\centering
\begin{tabular}{r|llll}
    & \textbf{Precision} & \textbf{Recall} & \textbf{TNR} & \textbf{Accuracy} \\ \hline
X1  & 0.7063                              & 1.0000                           & 0.0000                        & 0.7063                                                                                  \\
X2  & 0.0833                              & 1.0000                           & 0.2667                        & 0.3125                                                                                  \\
X3  & 0.5750                              & 0.9857                           & 0.4333                        & 0.6750                                                                                  \\
X4  & 0.4609                              & 1.0000                           & 0.3168                        & 0.5688                                                                                  \\
X5  & 0.5303                              & 0.9859                           & 0.3034                        & 0.6063                                                                                  \\
X6  & 0.4375                              & 1.0000                           & 0.0000                        & 0.4375                                                                                  \\
X7  & 0.6563                              & 1.0000                           & 0.0000                        & 0.6563                                                                                  \\
X8  & 0.3000                              & 1.0000                           & 0.3226                        & 0.4750                                                                                  \\
X9  & 0.1500                              & 1.0000                           & 0.2817                        & 0.3625                                                                                  \\
X10 & 0.5316                              & 1.0000                           & 0.0263                        & 0.5375                                                                                 
\end{tabular}
\label{LA}
\end{table}

\begin{table}[H]
\caption{The performance of distributed GLMM with 2-degree Gauss-Hermite transformation with significance threshold $\alpha=0.05$. TNR refers to True negative rate.}
\centering
\begin{tabular}{r|llll}
    & \textbf{Precision} & \textbf{Recall} & \textbf{TNR} & \textbf{Accuracy} \\ \hline
X1  & 0.5455                              & 1.0000                           & 0.0000                        & 0.5455                                                                                  \\
X2  & 0.3158                              & 1.0000                           & 0.3390                        & 0.4935                                                                                  \\
X3  & 0.6404                              & 1.0000                           & 0.4938                        & 0.7338                                                                                  \\
X4  & 0.6066                              & 1.0000                           & 0.4000                        & 0.6883                                                                                  \\
X5  & 0.5952                              & 1.0000                           & 0.3544                        & 0.6688                                                                                  \\
X6  & 0.5260                              & 1.0000                           & 0.0000                        & 0.5260                                                                                  \\
X7  & 0.6948                              & 1.0000                           & 0.0000                        & 0.6948                                                                                  \\
X8  & 0.5000                              & 0.9828                           & 0.4063                        & 0.6234                                                                                  \\
X9  & 0.4386                              & 1.0000                           & 0.3846                        & 0.5844                                                                                  \\
X10 & 0.5658                              & 1.0000                           & 0.0294                        & 0.5714      
\end{tabular}
\label{GH}
\end{table}

\begin{table}[H]
\caption{The result of centralized GLMM in R package}
\centering
\begin{tabular}{r|llllll}
\multicolumn{1}{l|}{}    & \textbf{Coef} & \textbf{Std.Err} & \textbf{z} & \textbf{P-value} & \textbf{{[}0.025} & \textbf{0.975{]}} \\ \hline
(Intercept)             & -5.882        & 0.133            & -44.294    & 0.000            & -6.142            & -5.621            \\
age                     & 0.043         & 0.001            & 30.918     & 0.000            & 0.040             & 0.045             \\
Gen\_M                  & 0.370         & 0.033            & 11.052     & 0.000            & 0.304             & 0.435             \\
race\_Asian             & 0.365         & 0.122            & 2.995      & 0.003            & 0.126             & 0.604             \\
race\_Caucasian         & 0.157         & 0.049            & 3.185      & 0.001            & 0.061             & 0.254             \\
race\_Other.Unknown     & 0.385         & 0.070            & 5.530      & 0.000            & 0.249             & 0.521             \\
ethnicity\_Not.Hispanic & -0.181        & 0.060            & -3.030     & 0.002            & -0.299            & -0.064            \\
ethnicity\_Unknown      & -0.088        & 0.069            & -1.274     & 0.203            & -0.224            & 0.048             \\
COPD\_Y                 & 0.096         & 0.038            & 2.556      & 0.011            & 0.022             & 0.169             \\
CHF\_Y                  & 0.153         & 0.040            & 3.796      & 0.000            & 0.074             & 0.232             \\
CKD\_Y                  & 0.029         & 0.044            & 0.674      & 0.500            & -0.056            & 0.115             \\
MS\_Y                   & -0.090        & 0.124            & -0.725     & 0.469            & -0.332            & 0.153             \\
RA\_Y                   & 0.144         & 0.070            & 2.048      & 0.041            & 0.006             & 0.281             \\
LU\_Y                   & -0.001        & 0.204            & -0.004     & 0.997            & -0.400            & 0.398             \\
HTN\_Y                  & 0.113         & 0.053            & 2.121      & 0.034            & 0.009             & 0.217             \\
IHD\_Y                  & 0.334         & 0.037            & 9.077      & 0.000            & 0.262             & 0.406             \\
DIAB\_Y                 & 0.149         & 0.035            & 4.214      & 0.000            & 0.080             & 0.218             \\
ASTH\_Y                 & -0.170        & 0.054            & -3.143     & 0.002            & -0.276            & -0.064            \\
Obese\_Y                & 0.211         & 0.043            & 4.893      & 0.000            & 0.126             & 0.295                   \end{tabular}
\label{r_result}
\end{table}

\begin{table}[H]
\caption{The result of federated GLMM with GH method}
\centering
\begin{tabular}{r|llllll}
\multicolumn{1}{l|}{}    & \textbf{Coef} & \textbf{Std.Err} & \textbf{z} & \textbf{P-value} & \textbf{{[}0.025} & \textbf{0.975{]}} \\ \hline
(Intercept)             & -3.476        & 0.066            & -53.064    & 0.000            & -3.605            & -3.348            \\
age                     & 0.043         & 0.001            & 55.794     & 0.000            & 0.041             & 0.044             \\
Gen\_M                  & 0.370         & 0.021            & 17.847     & 0.000            & 0.329             & 0.410             \\
race\_Asian             & 0.364         & 0.075            & 4.847      & 0.000            & 0.217             & 0.511             \\
race\_Caucasian         & 0.156         & 0.028            & 5.538      & 0.000            & 0.101             & 0.211             \\
race\_Other.Unknown     & 0.383         & 0.041            & 9.365      & 0.000            & 0.303             & 0.463             \\
ethnicity\_Not.Hispanic & -0.178        & 0.036            & -4.936     & 0.000            & -0.249            & -0.107            \\
ethnicity\_Unknown      & -0.091        & 0.042            & -2.154     & 0.031            & -0.174            & -0.008            \\
COPD\_Y                 & 0.096         & 0.026            & 3.700      & 0.000            & 0.045             & 0.146             \\
CHF\_Y                  & 0.154         & 0.029            & 5.205      & 0.000            & 0.096             & 0.211             \\
CKD\_Y                  & 0.028         & 0.031            & 0.901      & 0.367            & -0.033            & 0.090             \\
MS\_Y                   & -0.093        & 0.078            & -1.205     & 0.228            & -0.245            & 0.059             \\
RA\_Y                   & 0.144         & 0.049            & 2.959      & 0.003            & 0.048             & 0.239             \\
LU\_Y                   & 0.001         & 0.119            & 0.008      & 0.994            & -0.233            & 0.235             \\
HTN\_Y                  & 0.113         & 0.028            & 3.988      & 0.000            & 0.057             & 0.169             \\
IHD\_Y                  & 0.334         & 0.024            & 14.162     & 0.000            & 0.288             & 0.381             \\
DIAB\_Y                 & 0.149         & 0.022            & 6.677      & 0.000            & 0.105             & 0.192             \\
ASTH\_Y                 & -0.169        & 0.032            & -5.242     & 0.000            & -0.233            & -0.106            \\
Obese\_Y                & 0.211         & 0.028            & 7.483      & 0.000            & 0.156             & 0.267                 
\end{tabular}
\label{gh_result}
\end{table}

\begin{table}[H]
\caption{The result of federated GLMM with LA method}
\centering
\begin{tabular}{r|llllll}
\multicolumn{1}{l|}{}   & \textbf{Coef} & \textbf{Std.Err} & \textbf{z} & \textbf{P-value} & \textbf{{[}0.025} & \textbf{0.975{]}} \\
(Intercept)             & -3.162        & 0.064            & -49.437    & 0.000            & -3.288            & -3.037            \\
age                     & 0.041         & 0.001            & 54.895     & 0.000            & 0.040             & 0.043             \\
Gen\_M                  & 0.359         & 0.021            & 17.374     & 0.000            & 0.318             & 0.399             \\
race\_Asian             & 0.315         & 0.075            & 4.214      & 0.000            & 0.168             & 0.461             \\
race\_Caucasian         & 0.130         & 0.028            & 4.675      & 0.000            & 0.076             & 0.185             \\
race\_Other.Unknown     & 0.330         & 0.041            & 8.138      & 0.000            & 0.251             & 0.410             \\
ethnicity\_Not.Hispanic & -0.211        & 0.036            & -5.886     & 0.000            & -0.281            & -0.140            \\
ethnicity\_Unknown      & -0.114        & 0.042            & -2.707     & 0.007            & -0.197            & -0.031            \\
COPD\_Y                 & 0.098         & 0.026            & 3.750      & 0.000            & 0.047             & 0.149             \\
CHF\_Y                  & 0.156         & 0.030            & 5.234      & 0.000            & 0.098             & 0.215             \\
CKD\_Y                  & 0.029         & 0.032            & 0.928      & 0.353            & -0.033            & 0.092             \\
MS\_Y                   & -0.092        & 0.077            & -1.192     & 0.233            & -0.244            & 0.059             \\
RA\_Y                   & 0.138         & 0.049            & 2.829      & 0.005            & 0.042             & 0.234             \\
LU\_Y                   & -0.007        & 0.118            & -0.061     & 0.952            & -0.239            & 0.225             \\
HTN\_Y                  & 0.104         & 0.028            & 3.722      & 0.000            & 0.049             & 0.159             \\
IHD\_Y                  & 0.337         & 0.024            & 14.208     & 0.000            & 0.290             & 0.383             \\
DIAB\_Y                 & 0.144         & 0.022            & 6.461      & 0.000            & 0.100             & 0.187             \\
ASTH\_Y                 & -0.175        & 0.032            & -5.445     & 0.000            & -0.238            & -0.112            \\
Obese\_Y                & 0.182         & 0.028            & 6.450      & 0.000            & 0.127             & 0.237                  
\end{tabular}
\label{la_result}
\end{table}






\end{document}